\documentclass{article}
\usepackage{spconf,amsmath,graphicx}
\usepackage{multirow}
\usepackage{booktabs}
\usepackage{subcaption}
\usepackage{url}
\usepackage{hyperref}
\hypersetup{colorlinks=True, urlcolor=black}

\usepackage[textfont=it,tableposition=top]{caption}
\usepackage{color}

\title{Resource-Efficient Transfer Learning From Speech Foundation Model Using Hierarchical Feature Fusion}
\name{
\parbox{.95\linewidth}{\centering}}

\name{Zhouyuan Huo, Khe Chai Sim, Bo Li, Dongseong Hwang, Tara N. Sainath, Trevor Strohman}
\address{Google LLC, USA \\
\fontsize{9}{9}\selectfont\ttfamily\upshape
\{zhhuo,khechai\}@google.com}

\begin{document}
\ninept
\maketitle
\begin{abstract}
Self-supervised pre-training of a speech foundation model, followed by supervised fine-tuning, has shown impressive quality improvements on automatic speech recognition (ASR) tasks. Fine-tuning separate foundation models for many downstream tasks are expensive since the foundation model is usually very big. Parameter-efficient fine-tuning methods (e.g. adapter, sparse update methods) offer an alternative paradigm where a small set of parameters are updated to adapt the foundation model to new tasks. However, these methods still suffer from a high computational memory cost and slow training speed because they require backpropagation through the entire neural network at each step. In the paper, we analyze the performance of features at different layers of a foundation model on the speech recognition task and  propose a novel hierarchical feature fusion method for resource-efficient transfer learning from speech foundation models. Experimental results show that the proposed method can achieve better performance on speech recognition task than existing algorithms with fewer number of trainable parameters, less computational memory cost and faster training speed. After combining with Adapters at all layers, the proposed method can achieve the same performance as fine-tuning the whole model with $97\%$ fewer trainable encoder parameters and $53\%$ faster training speed.
\end{abstract}

\section{Introduction}
\label{sec:intro}

A foundation model~\cite{bommasani2021opportunities} is usually a big model trained on broad data (generally using self-supervision at scale) that can be fine-tuned to a wide range of downstream tasks and has aroused extensive attention due to its impressive quality improvements and emergent capabilities~\cite{brown2020language,radford2021learning,raffel2020exploring,liu2019roberta}.
In speech community, self-supervised pre-training speech foundation models on a large amount of unsupervised speech has shown impressive quality improvements on various speech recognition tasks \cite{zhang2022bigssl,hwang2022large}. There are two main categories of speech self-supervised learning algorithms. One direction is to reconstruct (APC~\cite{chung2020generative}, MPC~\cite{wang2020unsupervised}) or predict (Wav2vec~\cite{oord2018representation,Schneider2019_wav2vec,huo2021incremental}) the input feature directly. The other direction is building a BERT-style self-supervised learning model by bridging the gap between continuous speech signal and discrete text tokens,
such as Wav2vec 2.0~\cite{baevski2020wav2vec}, HuBERT~\cite{hsu2021hubert}, w2v-BERT~\cite{chung2021w2v}  and BEST-RQ~\cite{chiu2022self}. After pre-training the speech foundation model using the self-supervised loss, we initialize the encoder of the downstream task using the pre-trained model and fine-tune it on the supervised data.  

A large general-purpose foundation model with millions or even billions of parameters can be adapted to many downstream tasks. However, it is challenging to perform  separate adaptations for many tasks efficiently with only a small amount of supervised data each task. There have been existing works investigating to reduce the number of parameters required for fine-tuning the foundation model. BitFit~\cite{devlin2018bert} proposes a sparse-finetuning method where only the bias terms of the foundation model are updated.  Houlsby et al.~\cite{houlsby2019parameter} propose to insert Adapter modules between the layers in the fixed pre-trained model and each  module is a small trainable feed-forward neural network. Other works~\cite{karimi2021compacter,hu2021lora} reduce the number of parameters further by exploiting a low-rank approximation of the Adapter. Although these  parameter-efficient methods achieve decent performance on the downstream task with a significant reduction in the trainable parameters, their required computational memory cost and training time are still very high because of the following two reasons: 1) using the output of the highest layer in the foundation model only for downstream tasks, which leads to the inefficiency of the feature usage and requires to update the foundation model to adapt it to the downstream tasks;  2) adding/updating sparse parameters in the foundation model, which requires a full backpropagation process from the top to the bottom of the network to compute the gradients of the trainable parameters. Thus, a resource-efficient transfer learning method, which can achieve comparable performance with small number of trainable parameters, low computational memory cost and fast training speed, is required for efficient adaptation of the foundation model to many downstream tasks. 

Recently, Pasad et al.~\cite{pasad2021layer} analyze the layer-wise features of a self-supervised (wav2vec2.0) pre-trained speech representation model and finds that the middle layers encode the most contextual and high-level information.  The bottom or top few layers, on the other hand, focus on the lower-level information and encode more local representations. Arunkumar et al.~\cite{arunkumar2022investigation} investigate the ensemble features of self-supervised pre-trained models for
ASR and finds that features from different self-supervised learning methods are complementary and the ensemble of features is beneficial for the downstream speech recognition tasks. Although behaviors of the layer-wise features and features from multiple self-supervised pre-trained models are explored, neither of them consider the resource efficiency in the fine-tuning stage and there is no investigation about the feature fusion of layer-wise features from a single pre-trained model on downstream tasks. 

In this paper, we propose a novel resource-efficient transfer learning method for speech foundation models. specifically, we treat the foundation model as a frozen feature extractor and fuse the multi-level features from the foundation model hierarchically. We conduct extensive experiments to investigate different ways of feature fusion for the foundation model. Experimental results show that the proposed method can achieve better performance on the ASR task than existing parameter-efficient fine-tuning algorithms with fewer number of trainable parameters, less computational memory cost and faster training speed. After combining with Adapters at all layers, the proposed method can achieve the same performance as fine-tuning the whole model with $97\%$ fewer trainable encoder parameters and $53\%$ faster training speed.

\section{Experimental Setup}
\label{sec:experiments}

\subsection{Foundation Model And Task}
The foundation model used in the paper is a $2$-layer convolutional network followed by a $24$-layer conformer encoder with hidden dimension $1024$ and about $600$M parameters in total. Each conformer layer~\cite{gulati2020conformer} is a convolution-augmented
transformer network, which consists of attention, feed-forward and convolutional modules. The model input is a vector of
size $128$ logMel features and SpecAugment~\cite{Park2019} is also applied to increase model robustness. We pre-train the $600$M conformer encoder using the BEST-RQ \cite{chiu2022self} algorithm for $800$K steps. For the downstream speech recognition task, we initialize the encoder using the pre-trained speech foundation model and the output of the encoder is used as input to an RNN-T~\cite{zhang2022bigssl} along with a $6$-layer LSTM decoder and dimension $768$. We train with Adam optimizer for both pre-training and fine-tuning, and use exponential moving
averaging (EMA) with decay rate 0.9999 for fine-tuning only. 
We update the trainable encoder parameters and LSTM decoder which has $124$M trainable parameters on Voice Search data for $100$K steps. If not described explicitly, the parameter efficiency refers to the reduction of the trainable parameters in the encoder only. All experiments are performed on TPUs.

\begin{figure}[t]
    \centering
    \includegraphics[width=0.4\textwidth]{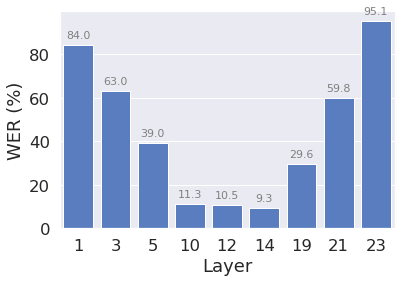}
    \caption{Voice Search WER when extracting features from different layers of the foundation model.}
    \label{fig:linear_wer}
\end{figure}

\subsection{Training Data} 

We use two sources of training data in this work. Following ~\cite{zhang2022bigssl}, we collect $800$K hours unsupervised  English Youtube data and pre-train the $600$M foundation model on the randomly segmented audio-only Youtube speech using the BEST-RQ algorithm~\cite{chiu2022self}.
In addition, the supervised English Voice Search (VS) data contains $5$K hours of labeled voice search audio~\cite{Narayanan2018} and is used to fine-tune the conformer encoder and RNNT-T decoder for the ASR task. All data are collected and deidentified in accordance with Google AI principles~\cite{aiprinciple}. 

\subsection{Evaluation}
In this paper, we calculate the word error rate (WER) on the Voice Search (VS) test dataset to measure the quality of the model on the downstream speech recognition task. Apart from WER, we compare the number of trainable parameters, computational memory cost and training speed at the same time for resource efficiency. The target of this paper is to propose a method, which can achieve low WER with small number of trainable parameters, low computational memory cost and fast training speed. 

\section{Linear Feature Fusion of The Foundation Model}

\label{sec:linear}
\begin{figure}[t]
    \centering
    \includegraphics[width=0.48\textwidth]{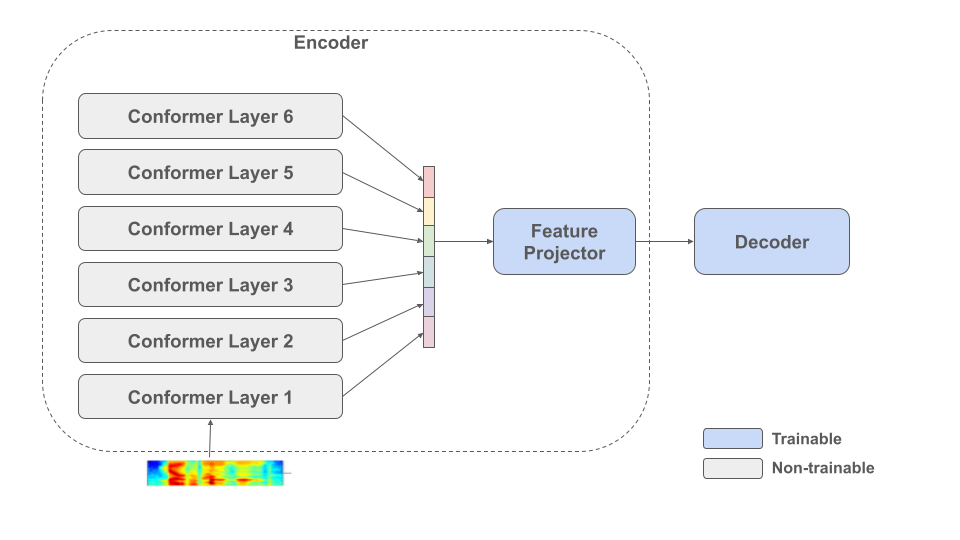}
    \caption{Linear feature fusion from multiple layers of the foundation model, using a $6$-layer conformer encoder as an example.}
    \label{fig:linear_alg}
\end{figure}

Previous parameter-efficient fine-tuning methods update the sparse parameters in the foundation model and use the output of the highest encoder layer only as the input to the RNN-T decoder, while the outputs of the intermediate layers are dropped after the forward pass. The proposed feature fusion method treat the foundation model as a frozen feature extractor and fuse the multi-level features from different layers linearly or hierarchically. Because there is no need to perform backward pass in the foundation model and only a few parameters are added on top of the outputs of the intermediate layers, the proposed feature fusion method is parameter-efficient and computation-efficient.  

\subsection{Performance of Single Layer Features}
\label{sec:linear_1}
To study the performance of the features from different layers of the foundation model, we extract outputs from layers \\
$\{1, 3, 5, 10, 12, 14, 19, 21, 23\}$ respectively and update the $124$M 6-layer LSTM decoder only on the Voice Search data. Figure \ref{fig:linear_wer} shows the WER of the corresponding layers and results present that models using features from middle layers perform better on the speech recognition task than features from bottom or top layers. This observation is consistent with ~\cite{pasad2021layer} that middle layers encode more contextual and high-level information which is more helpful for the speech recognition task than bottom or top layers. 

\begin{figure*}[t]
     \centering
     \begin{subfigure}[b]{0.48\textwidth}
         \centering
         \includegraphics[width=\textwidth]{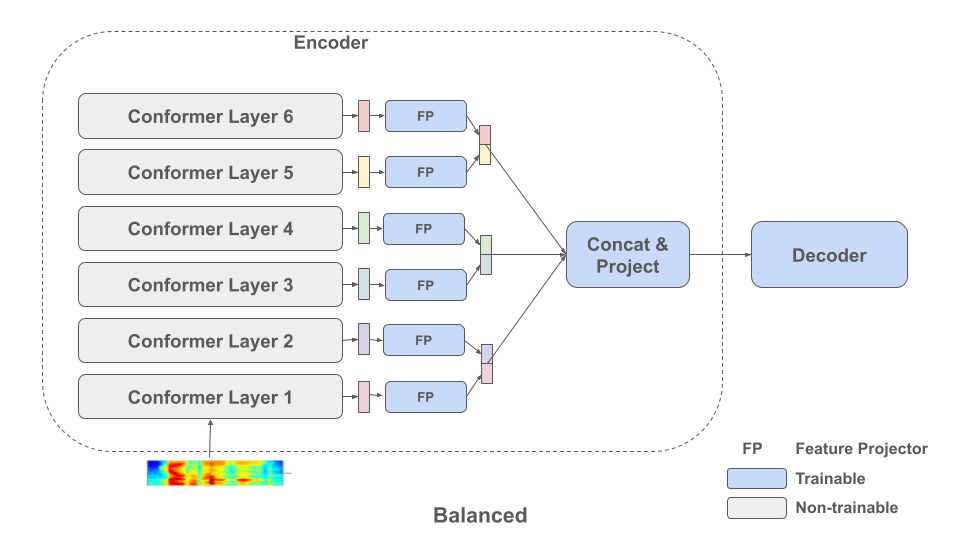}
     \end{subfigure}
     \begin{subfigure}[b]{0.48\textwidth}
         \centering
         \includegraphics[width=\textwidth]{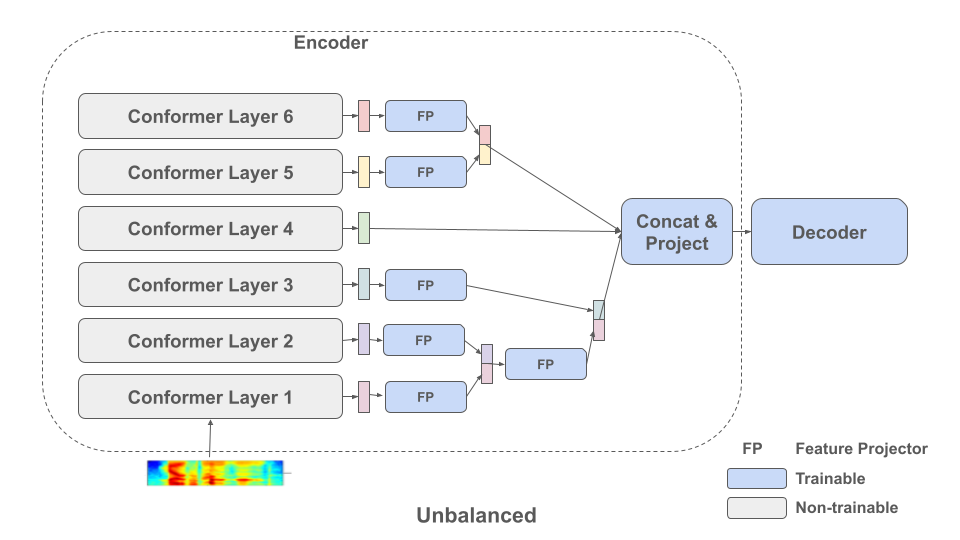}
     \end{subfigure}
        \caption{Balanced and unbalanced hierarchical feature fusion methods of the foundation model, using $6$-layer conformer encoder as an example. FP denotes a $1$-layer fully-connected network.}
    \label{fig:hff_alg}
\end{figure*}

\subsection{Linear Feature Fusion From Multiple Layers}
From Section \ref{sec:linear_1}, we know that features from different layers show different performance on the downstream speech recognition task. To investigate whether these features are complementary, we propose a linear feature fusion method and combine features from different layers linearly. As in Figure \ref{fig:linear_alg}, we firstly concatenate the features from multiple layers and project the concatenated feature to the required dimension using a fully-connected neural network. The decoder receives the output of the projector as input. All the conformer layers in the encoder are fixed while we update feature projector and decoder only using the RNN-T loss. 

\begin{table}[h]
    \centering
    \caption{Fusing features from multiple layers of the foundation model. Feature projector is a  $1$-layer fully-connected network for all combinations. }
    \label{tab:linear_multiple}
    \begin{tabular}{ccc}
        \toprule
        \textbf{Layer index} & \textbf{\# Parameters} & \textbf{VS WER} \\
        & \textbf{In Feature Projector} & {($\%$)} \\
        \midrule
         $11$ & $0.6$ M  & $11.2$ \\
        \midrule
         $23$ & $0.6$ M  & $91.9$ \\
        \midrule
         $11, 23$ & $1.3$ M  & $10.8$ \\
        \midrule
         $5, 11, 17, 23$ & $2.6$ M  & $9.3$ \\
        \midrule
         $2, 5, 8, 11, 14$ & $5.2$ M  & $8.1$ \\
         $ 17, 20, 23$ &  & \\
        \midrule
         $1, 3, 5, 7, 9, 11, 13$ & $7.9$ M  & $\textbf{8.0}$ \\
         $ 15, 17, 19, 21, 23$ & & \\
        \bottomrule
    \end{tabular}
\end{table}

\begin{figure}[h]
    \centering
    \includegraphics[width=0.45\textwidth]{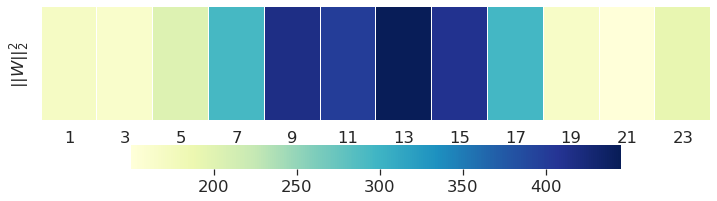}
    \caption{$\ell_2$ norm of the learned weight of each layer when fusing features from $12$ layers.}
    \label{fig:linear_weight}
\end{figure}

We compare the  Voice Search (VS) WER when fusing features from $\{1, 2, 4, 8, 12\}$ layers. Results in Table \ref{tab:linear_multiple} demonstrate the benefit of feature fusion from multiple layers. When fusing features from $12$ layers, we obtain the best VS WER $8.0\%$ with additional $7.9$M parameters in the feature projector. Figure \ref{fig:linear_weight} shows the norm of the learned weights for each layer when fusing $12$ features. The figure presents a higher weight for the middle layers and a lower weight for bottom or top layers. The results demonstrate that features from middle layers contribute more to the speech recognition task and adding features from bottom or top layers is also helpful.

\subsection{Increasing Depth of The Feature Projector}
We also explore to learn non-linear feature fusion by increasing the depth of the feature projector in Figure \ref{fig:linear_alg}. In Table \ref{tab:linear_depth}, we increase the depth of the fully-connected network from $1$ to $4$ layers with ReLU  activation while extracting features from the same $12$ layers used in the previous experiments, which was found to give the best results. Experimental results show that the model gets a better WER with a deeper feature projector and the VS WER becomes saturated at about $7.4\%$ after adding up to $3$ fully-connected layers. 

\begin{table}[h]
    \centering
    \caption{Increasing depth of the feature projector. Fusing features from $12$ layers as it gives the best results. }
    \label{tab:linear_depth}
    \begin{tabular}{ccc}
        \toprule
        \textbf{\# Layers} & \textbf{\# Parameters} & \textbf{VS WER} \\
        & \textbf{In Feature Projector} & {($\%$)} \\
        \midrule
         $1$ & $7.9$ M  & $8.0$ \\
        \midrule
         $2$ & $8.3$ M  & $7.5$ \\
        \midrule
         $3$ & $8.7$ M  & $\textbf{7.4}$ \\
        \midrule
         $4$ & $9.1$ M  & $\textbf{7.4}$ \\
        \bottomrule
    \end{tabular}
\end{table}

\section{Hierarchical Feature Fusion of the Foundation Model}
\label{sec:hff}
Knowing that features from different layers encode different levels of information, we also explore to fuse features in a hierarchical way rather than linearly. In this section, we propose a hierarchical feature fusion method and compare it with other parameter-efficient fine-tuning algorithms.

\begin{table*}[t]
    \centering
    \caption{Comparison with baselines and prameter-efficient methods. $\downarrow$ denotes the smaller the better. Second column shows the number of trainable parameters in the encoder only for the corresponding compared method and the whole $124$M LSTM decoder are trainable as well.}
    \label{tab:comparison}
    \begin{tabular}{ccccc}
        \toprule
        \textbf{Methods} & \textbf{\# Trainable } & \textbf{Computational } & \textbf{Training Speed } & \textbf{VS WER } \\
        & \textbf{Encoder Params $\downarrow$} & \textbf{Memory Cost $\downarrow$}& \textbf{Examples/Sec $\uparrow$} & \textbf{ ($\%) \downarrow$} \\
        \midrule
         Fine-tune  all &  $606.6$ M  & $13567$ MB & $1270$ & $5.5$  \\
        \midrule
         Fine-tune the highest encoder layer (FTHS) &  $25.4$ M  & $7563$ MB & $3616$ & $15.8$   \\
        \midrule
         BitFit & $0.1$ M  & $12443$ MB & $2824$ &  $6.5$ \\
        \midrule
         Adapter(d=$128$) at all layers & $6.4$ M & $12411$ MB & $2810$ & $6.4$  \\
        \midrule
         Adapter(d=$256$) at all layers & $13.3$ M & $12455$ MB & $2802$ & $6.1$  \\
        \midrule
         Adapter(d=$512$) at all layers & $25.9$ M & $12486$ MB & $2788$ & $6.1$  \\
        \midrule
         Adapter(d=$128$) at layers &&&& \\
         $\{13, 15, 17, 19, 21, 23\}$ & $2.3$ M  & $9340$ MB & $3251$ & $7.9$ \\
        \midrule
        \midrule
        Linear Feature Fusion & $8.7$ M & $7573$ MB & $3610$ & $7.4$ \\
        \midrule
        HFF-b & $12.3$ M  & $7648$ MB &  $3655$ & $7.0$ \\
        \midrule
         HFF-b + Adapter(d=$128$) at layers & & & \\
          $\{13, 15, 17, 19, 21, 23\}$   & $13.9$ M & $\textbf{9653}$ MB & $\textbf{3213}$ & $\textbf{6.0}$ \\
        \midrule
         HFF-b + Adapter(d=$128$) at all layers  & $18.6$ M & $12378$ MB & $\textbf{2750}$ & $\textbf{5.5}$ \\
        \bottomrule
    \end{tabular}
\end{table*}

\subsection{Hierarchical Feature Fusion From Multiple Layers}
\label{sec:hff_methods}

As in Figure \ref{fig:hff_alg}, we compare two hierarchical feature fusion methods (balanced and unbalanced) for the speech foundation model. For the balanced feature fusion method (HFF-b), we project and concatenate the neighboring pair-wise features, treating all layers equally. For the unbalanced feature fusion method (HFF-ub), on the other hand, we project and concatenate the neighboring features from bottom to the middle and from top to the middle. The intuition is that the middle layers encode high-level information while the bottom or top layers encode low-level information, such that more encoding is required for the features from these layers.

\begin{table}[h]
    \centering
    \caption{Comparison between balanced and unbalanced hierarchical feature fusion methods. Fusing features from $12$ layers.
    }
    \label{tab:hff_comp}
    \begin{tabular}{ccc}
        \toprule
        \textbf{Methods} & \textbf{\# Parameters} & \textbf{VS WER} \\
        & \textbf{In Feature Projector} &  \\
        \midrule
         HFF-b &  $12.3$ M  & $\textbf{7.0}$ \\
        \midrule
         HFF-ub & $12.3$  M  & $7.2$ \\
        \bottomrule
    \end{tabular}
\end{table}
We use a $1$-layer fully-connected network as FP in Figure \ref{fig:hff_alg} and the projector in the ``Concat $\&$ Project" is a $3$-layer fully-connected network. The FP projects a $1024$-d feature to $512$-d, such that the feature dimension remains unchanged after concatenation.  Table \ref{tab:hff_comp} shows that both methods achieve better VS WER than linear feature fusion, and HFF-b performs better than the HFF-ub on the speech recognition task with the same amount of parameters in the feature projector. Therefore,  we use balanced  hierarchical feature fusion (HFF-b) in the following experiments.

\subsection{Comparison with Parameter-Efficient Fine-Tuning Methods}
\label{sec:hff-comp}

To validate the proposed hierarchical feature fusion method, we compare it to several related algorithms. Specifically, we compare with two representative and strong
parameter-efficient methods: BitFit~\cite{zaken2021bitfit} and Adapter~\cite{houlsby2019parameter}. Each adapter module is inserted after each conformer encoder layer and is a randomly initialized $2$-layer feed-forward network with the bottleneck dimension $d$ from $\{128, 256, 512\}$.  We also fine-tune the highest conformer encoder layer (FTHST) as a baseline, which is computation-efficient because no backpropagation is required for the lower encoder layers. The parameter-efficient methods are applied to fine-tune the $600$M conformer encoder only, and the whole randomly initialized $124$M LSTM decoder is also updated simultaneously. Because the LSTM decoders are the same for all compared methods, we only compare the number of trainable encoder parameters in the experiments regarding parameter efficiency. Although the best VS WER can be achieved if we fine-tune all parameters of the model, it costs too much computational memory $13567$MB and the training speed is very slow at $1270$ examples/sec. Results in Table \ref{tab:comparison} show that Adapter's performance is better than BitFit or FTHS, but gets stuck at $6.1$ VS WER even if increasing the bottleneck dimension from $128$ to $512$. However, the Adapter(d=128) at all layers's training speed is $22\%$ slower and computational memory cost is $64\%$ higher than FTHS. With a very similar computational memory cost and training speed to FTHS, HFF-b can improve VS WER from $15.8\%$ to $7.0\%$. Comparing with Adapter(d=128) at layers $\{13, 15, 17, 19, 21, 23\}$, HFF-b achieves better VS WER with $12\%$ faster training speed and $18\%$ lower computation memory cost.  Combining the HFF-b with Adapter(d=128) at layers $\{13, 15, 17, 19, 21, 23\}$, we can achieve better VS WER $6.0\%$ than all compared parameter-efficient methods with fewer number of trainable parameters, less computational memory cost and faster training speed. If combining the proposed HFF-b with Adapter($d=128$) at all layers, we can achieve the same WER as fine-tuning all parameters of the RNN-T model  with $97\%$ fewer trainable encoder parameters and $53\%$ faster training speed. 
\section{Conclusion}
\label{sec:conclusion}
In this paper, we analyze the behavior of features from different layers of the foundation model for speech recognition task and  propose a hierarchical feature fusion method for resource-efficient transfer learning from the speech foundation model. Extensive results demonstrate that it achieves promising performance on the speech recognition task with fewer trainable encoder parameters, less computational cost and faster training speed.



\bibliographystyle{IEEEbib}
\bibliography{refs}

\begin{thebibliography}{10}

\bibitem{bommasani2021opportunities}
R.~Bommasani, D.~A. Hudson, E.~Adeli, R.~Altman, S.~Arora, S.~von Arx, M.~S.
  Bernstein, J.~Bohg, A.~Bosselut, E.~Brunskill, et~al.,
\newblock ``On the opportunities and risks of foundation models,''
\newblock {\em arXiv preprint arXiv:2108.07258}, 2021.

\bibitem{brown2020language}
T.~Brown, B.~Mann, N.~Ryder, M.~Subbiah, J.~D. Kaplan, P.~Dhariwal,
  A.~Neelakantan, P.~Shyam, G.~Sastry, A.~Askell, et~al.,
\newblock ``Language models are few-shot learners,''
\newblock {\em Advances in neural information processing systems}, vol. 33, pp.
  1877--1901, 2020.

\bibitem{radford2021learning}
A.~Radford, J.~W. Kim, C.~Hallacy, A.~Ramesh, G.~Goh, S.~Agarwal, G.~Sastry,
  A.~Askell, P.~Mishkin, J.~Clark, et~al.,
\newblock ``Learning transferable visual models from natural language
  supervision,''
\newblock in {\em International Conference on Machine Learning}. PMLR, 2021,
  pp. 8748--8763.

\bibitem{raffel2020exploring}
C.~Raffel, N.~Shazeer, A.~Roberts, K.~Lee, S.~Narang, M.~Matena, Y.~Zhou,
  W.~Li, P.~J. Liu, et~al.,
\newblock ``Exploring the limits of transfer learning with a unified
  text-to-text transformer.,''
\newblock {\em J. Mach. Learn. Res.}, vol. 21, no. 140, pp. 1--67, 2020.

\bibitem{liu2019roberta}
Y.~Liu, M.~Ott, N.~Goyal, J.~Du, M.~Joshi, D.~Chen, O.~Levy, M.~Lewis,
  L.~Zettlemoyer, and V.~Stoyanov,
\newblock ``Roberta: A robustly optimized bert pretraining approach,''
\newblock {\em arXiv preprint arXiv:1907.11692}, 2019.

\bibitem{zhang2022bigssl}
Y.~Zhang, D.~S. Park, W.~Han, J.~Qin, A.~Gulati, J.~Shor, A.~Jansen, Y.~Xu,
  Y.~Huang, S.~Wang, et~al.,
\newblock ``Bigssl: Exploring the frontier of large-scale semi-supervised
  learning for automatic speech recognition,''
\newblock {\em IEEE Journal of Selected Topics in Signal Processing}, 2022.

\bibitem{hwang2022large}
D.~Hwang, A.~Misra, Z.~Huo, N.~Siddhartha, S.~Garg, D.~Qiu, K.~C. Sim,
  T.~Strohman, F.~Beaufays, and Y.~He,
\newblock ``Large-scale asr domain adaptation using self-and semi-supervised
  learning,''
\newblock in {\em ICASSP 2022-2022 IEEE International Conference on Acoustics,
  Speech and Signal Processing (ICASSP)}. IEEE, 2022, pp. 6627--6631.

\bibitem{chung2020generative}
Y.-A. Chung and J.~Glass,
\newblock ``Generative pre-training for speech with autoregressive predictive
  coding,''
\newblock in {\em ICASSP 2020-2020 IEEE International Conference on Acoustics,
  Speech and Signal Processing (ICASSP)}. IEEE, 2020, pp. 3497--3501.

\bibitem{wang2020unsupervised}
W.~Wang, Q.~Tang, and K.~Livescu,
\newblock ``Unsupervised pre-training of bidirectional speech encoders via
  masked reconstruction,''
\newblock in {\em ICASSP 2020-2020 IEEE International Conference on Acoustics,
  Speech and Signal Processing (ICASSP)}. IEEE, 2020, pp. 6889--6893.

\bibitem{oord2018representation}
A.~v.~d. Oord, Y.~Li, and O.~Vinyals,
\newblock ``Representation learning with contrastive predictive coding,''
\newblock {\em arXiv preprint arXiv:1807.03748}, 2018.

\bibitem{Schneider2019_wav2vec}
S.~Schneider, A.~Baevski, R.~Collobert, and M.~Auli,
\newblock ``{wav2vec: Unsupervised Pre-Training for Speech Recognition},''
\newblock in {\em Proc. Interspeech 2019}, 2019, pp. 3465--3469.

\bibitem{huo2021incremental}
Z.~Huo, D.~Hwang, K.~C. Sim, S.~Garg, A.~Misra, N.~Siddhartha, T.~Strohman, and
  F.~Beaufays,
\newblock ``Incremental layer-wise self-supervised learning for efficient
  speech domain adaptation on device,''
\newblock {\em arXiv preprint arXiv:2110.00155}, 2021.

\bibitem{baevski2020wav2vec}
A.~Baevski, H.~Zhou, A.~Mohamed, and M.~Auli,
\newblock ``wav2vec 2.0: A framework for self-supervised learning of speech
  representations,''
\newblock {\em arXiv preprint arXiv:2006.11477}, 2020.

\bibitem{hsu2021hubert}
W.-N. Hsu, B.~Bolte, Y.-H.~H. Tsai, K.~Lakhotia, R.~Salakhutdinov, and
  A.~Mohamed,
\newblock ``Hubert: Self-supervised speech representation learning by masked
  prediction of hidden units,''
\newblock {\em IEEE/ACM Transactions on Audio, Speech, and Language
  Processing}, vol. 29, pp. 3451--3460, 2021.

\bibitem{chung2021w2v}
Y.-A. Chung, Y.~Zhang, W.~Han, C.-C. Chiu, J.~Qin, R.~Pang, and Y.~Wu,
\newblock ``W2v-bert: Combining contrastive learning and masked language
  modeling for self-supervised speech pre-training,''
\newblock in {\em 2021 IEEE Automatic Speech Recognition and Understanding
  Workshop (ASRU)}. IEEE, 2021, pp. 244--250.

\bibitem{chiu2022self}
C.-C. Chiu, J.~Qin, Y.~Zhang, J.~Yu, and Y.~Wu,
\newblock ``Self-supervised learning with random-projection quantizer for
  speech recognition,''
\newblock {\em arXiv preprint arXiv:2202.01855}, 2022.

\bibitem{devlin2018bert}
J.~Devlin, M.-W. Chang, K.~Lee, and K.~Toutanova,
\newblock ``Bert: Pre-training of deep bidirectional transformers for language
  understanding,''
\newblock {\em arXiv preprint arXiv:1810.04805}, 2018.

\bibitem{houlsby2019parameter}
N.~Houlsby, A.~Giurgiu, S.~Jastrzebski, B.~Morrone, Q.~De~Laroussilhe,
  A.~Gesmundo, M.~Attariyan, and S.~Gelly,
\newblock ``Parameter-efficient transfer learning for nlp,''
\newblock in {\em International Conference on Machine Learning}. PMLR, 2019,
  pp. 2790--2799.

\bibitem{karimi2021compacter}
R.~Karimi~Mahabadi, J.~Henderson, and S.~Ruder,
\newblock ``Compacter: Efficient low-rank hypercomplex adapter layers,''
\newblock {\em Advances in Neural Information Processing Systems}, vol. 34, pp.
  1022--1035, 2021.

\bibitem{hu2021lora}
E.~J. Hu, Y.~Shen, P.~Wallis, Z.~Allen-Zhu, Y.~Li, S.~Wang, L.~Wang, and
  W.~Chen,
\newblock ``Lora: Low-rank adaptation of large language models,''
\newblock {\em arXiv preprint arXiv:2106.09685}, 2021.

\bibitem{pasad2021layer}
A.~Pasad, J.-C. Chou, and K.~Livescu,
\newblock ``Layer-wise analysis of a self-supervised speech representation
  model,''
\newblock in {\em 2021 IEEE Automatic Speech Recognition and Understanding
  Workshop (ASRU)}. IEEE, 2021, pp. 914--921.

\bibitem{arunkumar2022investigation}
A.~Arunkumar, V.~N. Sukhadia, and S.~Umesh,
\newblock ``Investigation of ensemble features of self-supervised pretrained
  models for automatic speech recognition,''
\newblock {\em arXiv preprint arXiv:2206.05518}, 2022.

\bibitem{gulati2020conformer}
A.~Gulati, J.~Qin, C.-C. Chiu, N.~Parmar, Y.~Zhang, J.~Yu, W.~Han, S.~Wang,
  Z.~Zhang, Y.~Wu, et~al.,
\newblock ``Conformer: Convolution-augmented transformer for speech
  recognition,''
\newblock {\em arXiv preprint arXiv:2005.08100}, 2020.

\bibitem{Park2019}
D.~S. Park, W.~Chan, Y.~Zhang, C.-C. Chiu, B.~Zoph, E.~D. Cubuk, and Q.~V. Le,
\newblock ``{SpecAugment: A Simple Data Augmentation Method for Automatic
  Speech Recognition},''
\newblock in {\em Proc. Interspeech 2019}, 2019, pp. 2613--2617.

\bibitem{Narayanan2018}
A.~{Narayanan}, A.~{Misra}, K.~C. {Sim}, G.~{Pundak}, A.~{Tripathi},
  M.~{Elfeky}, P.~{Haghani}, T.~{Strohman}, and M.~{Bacchiani},
\newblock ``Toward domain-invariant speech recognition via large scale
  training,''
\newblock in {\em 2018 IEEE Spoken Language Technology Workshop (SLT)}, 2018,
  pp. 441--447.

\bibitem{aiprinciple}
``Artificial intelligence at google: Our principles,''
  \url{https://ai.google/principles/}.

\bibitem{zaken2021bitfit}
E.~B. Zaken, S.~Ravfogel, and Y.~Goldberg,
\newblock ``Bitfit: Simple parameter-efficient fine-tuning for
  transformer-based masked language-models,''
\newblock {\em arXiv preprint arXiv:2106.10199}, 2021.

\end{thebibliography}

\end{document}